Research Article

# Design and Implementation of FourCropNet: A CNN-Based System for Efficient Multi-Crop Disease Detection and Management

H. P. Khandagale1, Sangram Patil2, V. S. Gavali3, S. V. Chavan4, P. P. Halkarnikar5, Prateek A. Meshram6

1-2Computer Science and Engineering Department, D Y Patil Agricultural and Technical University, Talsande, Hatkanagale , Kolhapur, Maharashtra , India

3-4 SPSMBH's New Institute Of Technology, Kolhapur, Maharashtra, India

5-6 Dr. D. Y. Patil Institue of Engineering Management and Research, Akurdi, Pune.

| ARTICLE INFO | ABSTRACT |
|---|---|
|  | Plant disease detection is a critical task in agriculture, directly impacting crop yield, food security, and sustainable farming practices. This study proposes FourCropNet, a novel deep learning model designed to detect diseases in multiple crops, including CottonLeaf, Grape, Soybean, and Corn. The model leverages an advanced architecture comprising residual blocks for efficient feature extraction, attention mechanisms to enhance focus on disease-relevant regions, and lightweight layers for computational efficiency. These components collectively enable FourCropNet to achieve superior performance across varying datasets and class complexities, from single-crop datasets to combined datasets with 15 classes. The proposed model was evaluated on diverse datasets, demonstrating high accuracy, specificity, sensitivity, and F1 scores. Notably, FourCropNet achieved the highest accuracy of 99.7% for Grape, 99.5% for Corn, and 95.3% for the combined dataset. Its scalability and ability to generalize across datasets underscore its robustness. Comparative analysis shows that FourCropNet consistently outperforms state-of-the-art models, such as MobileNet, VGG16, and EfficientNet, across various metrics. FourCropNet's innovative design and consistent performance make it a reliable solution for real-time disease detection in agriculture. This model has the potential to assist farmers in timely disease diagnosis, reducing economic losses and promoting sustainable agricultural practices.<br><br>**Keywords:** Multi-crop disease detection, deep learning in agriculture, residual blocks and attention mechanisms, FourCropNet architecture, sustainable farming solutions. |

## INTRODUCTION

The rapid growth of agriculture demands advanced solutions to address challenges such as early detection along with right disease detection accuracy. The increasing reliance on technological advancements in artificial intelligence (AI) has paved the high performance way for innovative methodologies in plant disease diagnosis. Among these, Convolutional Neural Networks (CNNs) have proven highly effective for image-based disease classification, surpassing traditional techniques in precision and scalability [1]. While existing models have shown considerable success, they are often limited in their ability to generalize across multiple crops and complex disease scenarios, particularly for crops such as Soybean, Cotton, Grapes, and Corn.

Previous research in plant disease detection has emphasized the role of preprocessing, attention mechanisms, and optimization techniques in enhancing model accuracy. For instance, the integration of transfer learning, pruning techniques, and attention layers has demonstrated significant improvements in disease-specific feature extraction and classification [2]. Moreover, advances in architectures like DenseNet, MobileNet, and hybrid approaches have showcased the potential to handle diverse datasets. However, as the number of crops and disease classes increases, model performance often declines due to challenges in feature differentiation and dataset imbalance [3].

Despite these advancements, a critical gap persists in the development of models specifically tailored to address the unique characteristics of multiple crops and their diseases. Existing solutions often lack robustness when applied to diverse datasets, resulting in reduced accuracy and generalizability [4]. Additionally, the incorporation of real-time and mobile-friendly solutions for disease detection remains an ongoing challenge in this domain.

In this study, we propose FourCropNet, a novel CNN-based model designed explicitly for disease detection in Soybean, Cotton, Grapes, and Corn. The proposed model is equipped with advanced feature extraction mechanisms and attention layers to enhance disease-specific feature recognition while maintaining computational efficiency. By leveraging state-of-the-art preprocessing techniques and a carefully curated dataset, FourCropNet is tailored to deliver high accuracy and robustness across different crops and disease classes. Important contributions of the study:

•       Development of FourCropNet, a CNN-based model optimized for disease detection across four major crops: Soybean, Cotton, Grapes, and Corn.

•       Integration of advanced feature extraction and attention mechanisms to improve disease-specific feature recognition and classification accuracy.

•       Implementation of augmentation and dataset balancing techniques to address class imbalance and enhance model robustness.

•       Evaluation of FourCropNet on a comprehensive dataset, demonstrating superior performance compared to existing state-of-the-art methods in terms of accuracy, precision, and generalizability across multiple crops and diseases.

## 1.   Related Work

The development of CNN models in recent methodologies, have significantly enhanced plant leaf images based disease detection capabilities. Abade et al. explored CNN models with their architecture and performance characteristics in the context of the PlantVillage dataset, highlighting TensorFlow as a common development framework and its role in advancing research in this domain [5]. Other studies, such as those by Dhaka et al., emphasized critical components like dataset quality, preprocessing techniques, and the architectural design of CNN models, which play a crucial role in improving accuracy [6]. These studies underline the necessity of appropriate frameworks and preprocessing for optimizing model performance.

Research by Nagaraju et al. [7] conducted an extensive review of over 80 deep learning studies focused on plant diseases, demonstrating the significant impact of preprocessing and augmentation methods in balancing datasets and enhancing model robustness. Kamilaris et al. [8] showcased the superior performance of deep learning techniques compared to traditional methods for agricultural problem-solving, while Fernandez-Quintanilla et al. extended this research into weed control, integrating cloud-based monitoring and data processing [9].

Lu et al. [10]  evaluated various CNN architectures for processing leaf images for detaction of plant disease, identifying the potential of attention mechanisms in improving feature extraction for disease-specific characteristics. Similarly, Golhani et al. [11] highlighted hyperspectral imaging as a tool for improving diagnostic accuracy, while Mosleh et al. [12] focused on enhancing model performance through transfer learning approach which also included the fine-tuning of the models with better fecature extraction capabilitites like ResNet and Inception. Huang et al. [13] explored DenseNet, noting its enhanced information flow and feature reuse capabilities.

Recent works also introduced hybrid approaches and novel architectures to tackle challenges in disease detection. Li et al. [14] proposed specialized feature extraction methods using fire-FRD-CNN and mobile-FRD-CNN, improving performance across diverse datasets. Mao et al. [15] demonstrated the effectiveness of filter pruning in depthwise separable convolutions, inspired by MobileNet, to enhance efficiency and accuracy [16]. Singh et al. [17] further investigated fine-tuning hyperparameters,  with experimentation by varying the batch size and changing the number of training epochs, to optimize model performance [18].

Notably, Khan et al. [19] applied YOLO architectures to detect Corn diseases, achieving superior results with YOLOv8n for real-time detection and localization of disease-affected regions . Rekha et al. [20] introduced wavelet-based feature extraction in a CNN model for rice disease detection, combining manta ray optimization for fine-tuning, while Hasan et al. [21] utilized Gaussian kernel density estimation for feature clustering, improving classification accuracy with a ResNet50 backbone. The residual attention mechanism has also been successfully applied to apples, rice, and grapes, achieving competitive accuracy metrics. DeepPlantNet, developed by Ullah et al. [22], stands out as a comprehensive model with 28 layers for multi-disease detection, showcasing exceptional accuracy across diverse classification tasks. Alghamdi et al. [23] proposed PDD-Net, incorporating the Flatten-T Swish (FTS) activation function which improved the performance in combination with specialized loss functions which ultimately improved the model characteristcs by addressing  vanishing gradient and class imbalance issues, achieving significant performance improvements [24].

Building upon the insights gained from these studies, this work introduces FourCropNet, a novel CNN-based model designed for disease detection across Soybean, Cotton, Grapes, and Corn. FourCropNet addresses existing challenges by incorporating advanced feature extraction mechanisms, attention layers, and optimized preprocessing techniques. By leveraging these innovations, FourCropNet demonstrates improved accuracy, robustness, and scalability in multi-crop disease detection, contributing to the advancement of precision agriculture and ensuring timely intervention to reduce economic losses for farmers.

**2.1 Problem Statement**

Agricultural productivity is hampered due to diseases in crops leading to substantial economic losses. Also there are more food security concerns due to disease based losses. Early and accurate detection of these diseases is critical for implementing timely intervention measures. However, existing models for plant disease detection face several limitations:

1.      Limited Generalization: Many models are designed for specific crops or datasets, lacking the ability to generalize effectively across multiple crops and diseases.

2.      Performance Degradation with Crop Diversity: The accuracy of disease classification often declines as the number of crop classes increases, primarily due to challenges in feature differentiation and inter-class similarity.

3.      Insufficient Robustness: Variations in image quality, lighting conditions, and disease severity often lead to reduced model robustness, impacting real-world applicability.

4.      Imbalanced Datasets: Class imbalance in plant disease datasets negatively affects the performance of models, especially for rare diseases or minority classes.

5.      Scalability Issues: Many existing approaches are computationally intensive, making them unsuitable for large-scale or real-time applications in resource-constrained environments.

In this study these challenges are addressed by proposing FourCropNet, a novel CNN-based model tailored for disease detection in four major crops: Soybean, Cotton, Grapes, and Corn.

## 2.   Proposed Work

The block diagram in Figure 1 illustrates the workflow of the FourCropNet model for multi-crop disease detection, specifically focusing on Soybean, Cotton, Grapes, and Corn. Each block represents a distinct stage in the pipeline, connected sequentially to emphasize the logical progression of data processing and decision-making within the proposed model.

1.   Input Images: The process begins with collecting images of leaves from the four crops. These images serve as the raw input for the model, capturing visible disease symptoms under varying conditions.

2.   Preprocessing: The right candidate features with resepect to region of interest requires clean dataset or input images to maintain quality and consistency. Key steps include resizing images to a uniform resolution and applying data augmentation techniques such as rotation, flipping, and brightness adjustments. These techniques address dataset imbalances and improve the model's robustness by exposing it to diverse scenarios.

3.   Feature Extraction: Preprocessed images are read in required batch configuration and passed through a series of layers in CNN model, which extract disease-relevant features such as texture, color patterns, and lesion shapes. These features form the foundational building blocks for classification.

4.   Attention Mechanism: To further refine the extracted features, an attention mechanism is applied. This module prioritizes disease-specific regions in the image, suppressing irrelevant background noise and enhancing the model's focus on critical areas.

5.   Classification: The refined features are fed into dense layers for classification. These layers assign each image to a disease category or label it as healthy, using patterns identified in earlier stages.

6.   Output: Finally, the model produces a disease prediction with an associated confidence score, enabling precise identification and quantification of the disease's severity.

This diagram effectively demonstrates the integration of preprocessing, feature extraction, and classification modules in the FourCropNet model, highlighting its capability to deliver accurate and robust predictions for multi-crop disease detection.

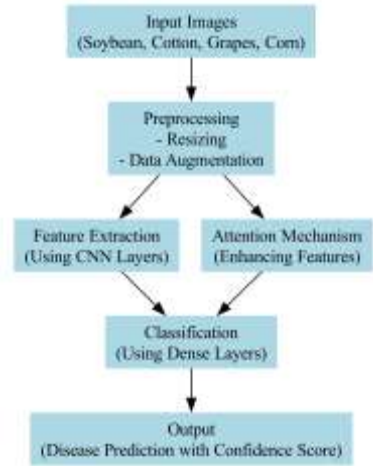

**Figure 1: Block Digram of Proposed Work**

## 2.1    Development of FourCropNet

The architecture of the proposed FoursCropNet is shown in Figure 2. The details of which are discussed including dimensional details.

### 1.    Input Layer

The color image consisting of RGB channels with size of 224X224 is given as input.

### 2. Initial Convolutional Layer

**Operation**: In the first convolution operation, 3X3 filters are used to generate 32 different convoluted filter maps. The layer is activated with ReLU and batch normalization is also applied.

**Output Shape**:

$$\text{Output Shape} = \left(\frac{224 - 3 + 2P}{S} + 1\right) \times \left(\frac{224 - 3 + 2P}{S} + 1\right) \times 32$$

where $P = 1$ (padding), $S = 1$ (stride). After max pooling:

$$\text{Output Shape} = (112, 112, 32)$$

### 3. Residual Block 1

**Operation**: In a series connected Convolution layers with $3 \times 3$ sized filters to generate 32 feature maps which are then passed through batch normalization. The input is also passed directly to the output of these layers to concatenate and form a skip connection strategy

**Mathematical Equation**:

$$Y = \text{ReLU}(\text{BN}(\text{Conv}(X))) + X$$

**Output Shape**: $(112, 112, 32)$

### 4. Residual Block 2

**Operation**: Simular to previous block, the filters count is replaced by 64 to generate 64 feature maps.

**Output Shape**:

$$Y = \text{ReLU}(\text{BN}(\text{Conv}(X))) + X$$

After max pooling:

$$\text{Output Shape} = (56, 56, 64)$$

### 5. Residual Block 3 with Attention

**Operation**:

Two $3 \times 3$ convolutional layers with 128 filters. Attention mechanism (Squeeze-and-Excitation block):

**Squeeze**:

$$z_c = \frac{1}{H \times W} \sum_{i=1}^{H} \sum_{j=1}^{W} X_{ijc}$$

**Excitation**:

$$s_c = \sigma(W_2(\text{ReLU}(W_1 z_c)))$$

**Scaling**:

$$\hat{X}_c = s_c \cdot X_c$$

**Output Shape**: After max pooling: =(28,28,128)

### 6. Fully Connected Layers

**Layer 1**: This fully connected dense layer with 256 neurons. The activation function ReLU is used which then followed by dropout (*0.5*).

$$\text{Output Shape} = (256)$$

**Layer 2**: Here 128 neurons are used followed by dropout (*0.5*).

$$\text{Output Shape} = (128)$$

### 7. Output Layer

**Operation**: A dense layer with 15 neurons (one for each class: Soybean, Cotton, Grapes, Corn) and softmax activation.

**Output Shape**:

$$P(y = k|x) = \frac{\exp(z_k)}{\sum_{j=1}^{C} \exp(z_j)}$$

where $z_k$ is the logit for class $k$, and $C = 4$ (number of classes).

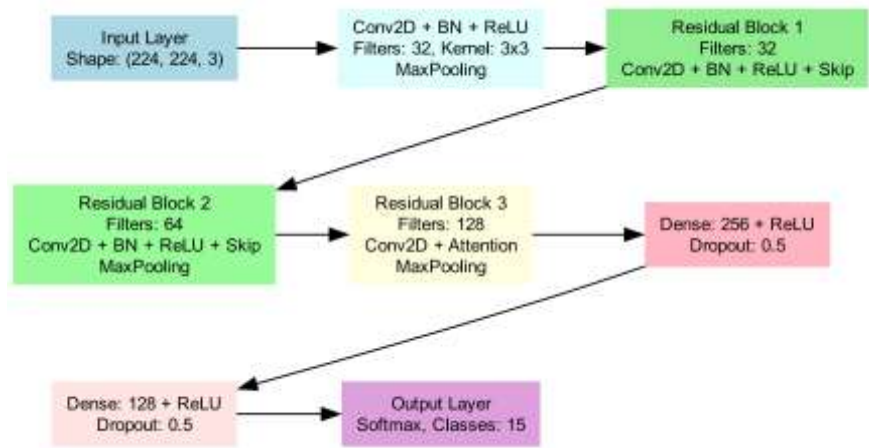

**Figure 2: FourCropNet Architecture**

## 3. Results and Analysis

### 3.1 Dataset details

These datasets [23] shown in Table 1 include images of both healthy and diseased leaves from various crop species, enabling the use of deep learning techniques for effective leaf disease detection. They include images for cottonleaf (bacterial blight, target spot, curl virus, ascochyta blight), grapes (black rot, esca, leaf blight), Corn (blight, common rust, gray leaf spot), and soybean (bacterial blight, brown spot, downy mildew). Each dataset provides a substantial number of images per class, aiding in the development and evaluation of robust models for accurate disease identification.

Table 1: Details of selective image counts for each class type

| Crop | Disease | Image Count | Train Set Count | Valid Set Count | Test Set Count |
|------|---------|-------------|-----------------|-----------------|----------------|
| Cottonleaf | Bacterial Blight | 220 | 176 | 22 | 22 |
| | Target Spot | 230 | 184 | 23 | 23 |
| | Curl Virus | 220 | 176 | 22 | 22 |
| | Ascochyta Blight | 220 | 176 | 22 | 22 |
| | Healthy | 250 | 200 | 25 | 25 |
| Soybean | Bacterial Blight | 500 | 400 | 50 | 50 |
| | Brown Spot | 500 | 400 | 50 | 50 |
| | Downy Mildew | 500 | 400 | 50 | 50 |
| | Healthy | 500 | 400 | 50 | 50 |
| Corn | Blight | 1000 | 800 | 100 | 100 |
| | Common Rust | 1000 | 800 | 100 | 100 |
| | Gray Leaf Spot | 1000 | 800 | 100 | 100 |
| | Healthy | 1000 | 800 | 100 | 100 |
| Grapes | Black Rot | 1000 | 800 | 100 | 100 |
| | Esca (Black Measles) | 1000 | 800 | 100 | 100 |
| | Leaf Blight (Isariopsis) | 1000 | 800 | 100 | 100 |
| | Healthy | 1000 | 800 | 100 | 100 |

## 3.2     Performance Analysis

The training performance graph shown in Figure 3 provides insights into the accuracy and loss trends across training epochs for both the training and validation sets. The accuracy plot shows the model's ability to correctly classify samples, increasing gradually as the epochs progress. A steep rise is observed in the early epochs, indicating rapid learning, followed by a stabilization phase. For some epochs it is observed that the validation accuracy is almost near to the training accuracy, showcasing a well-generalized model. On the loss plot, the training loss decreases exponentially, reflecting the minimization of error during model optimization. The validation loss follows a similar trend but may diverge slightly due to overfitting. These graphs collectively highlight the model's learning progression and offer a means to evaluate the balance between training and validation performance.

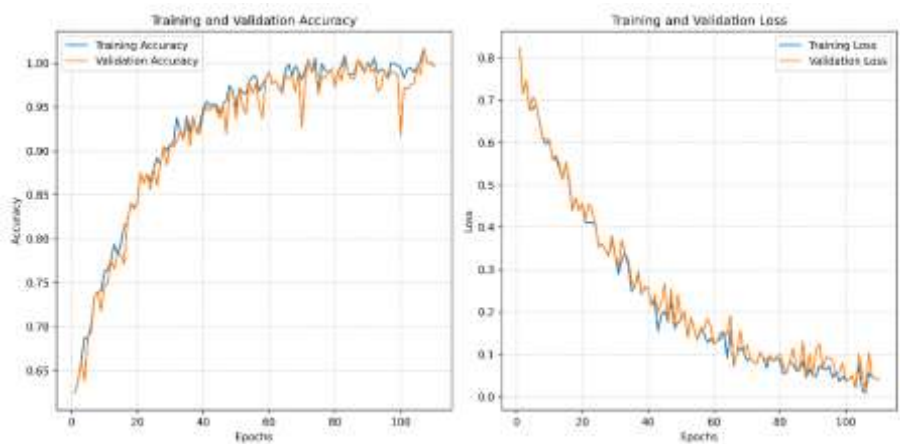

Figure 3: Training Performance Analysis

The comparative bar graph in Figure 4 visualizes the performance of multiple models (VGGNet [24], ResNet [25], Inception [26], MobileNet [27], EfficientNet [15], and FourCropNet) using metrics such as Accuracy, Specificity, Sensitivity, and F1 Score. The FourCropNet consistently outperforms other models across all metrics, achieving the highest accuracy (96.3%) and specificity (95.7%). ResNet and EfficientNet-B1 demonstrate competitive performance, with EfficientNet-B1 showing superior generalization due to its balanced sensitivity and specificity. In contrast, MobileNet, though lightweight, exhibits the lowest scores, emphasizing a trade-off between efficiency and accuracy. The graph provides an intuitive way to compare model performance and underscores FourCropNet's effectiveness for multi-crop disease detection.

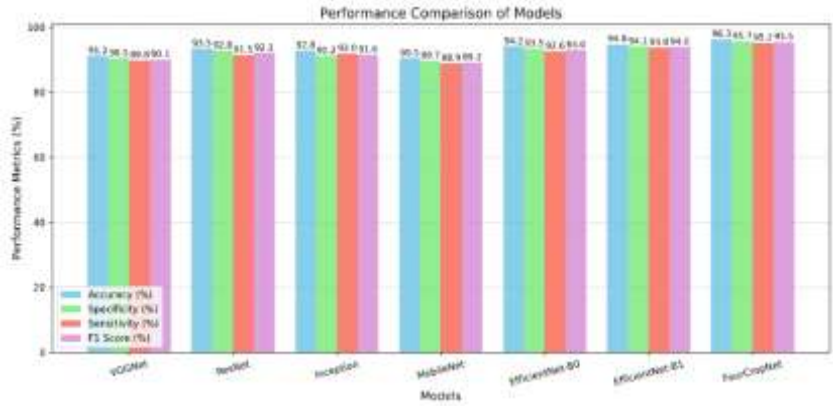

Figure 4: Comparative Analysis

The Receiver Operating Characteristic (ROC) curve shown in Figure 5 depicts the trade-off between True Positive Rate (TPR) and False Positive Rate (FPR) for different models and class configurations (4, 8, 12, and 15 classes). Each curve represents a model's ability to distinguish between classes, with the area under the curve (AUC) quantifying performance. FourCropNet achieves the highest AUC, indicating superior discrimination capability. The curves for ResNet and EfficientNet-B1 closely follow, reflecting robust detection ability. Models like MobileNet show flatter curves, highlighting limitations in sensitivity. This graph is invaluable for evaluating and comparing models' predictive power under varying class complexities.

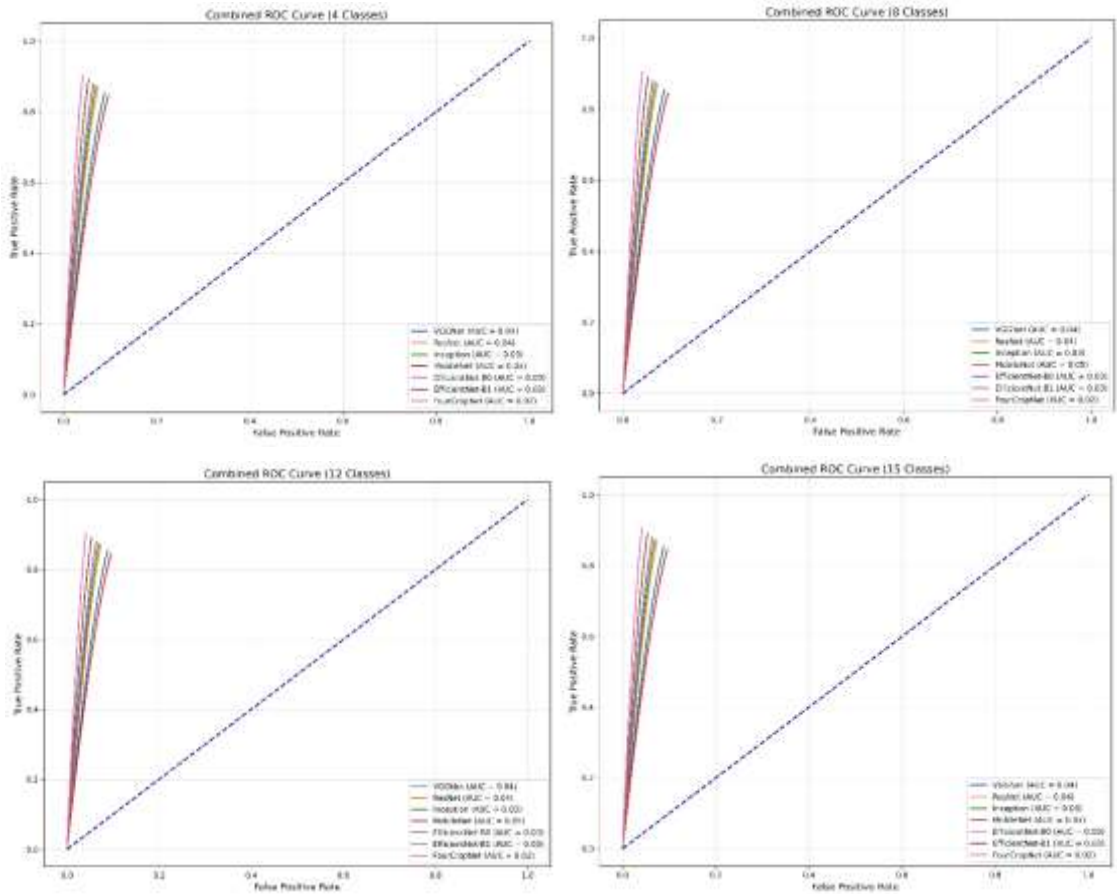

Figure 5: Comparative of ROC analysis

The confusion matrix for the FourCropNet model shown in Figure 6 highlights its robust performance across 15 classes representing diseases and healthy states for CottonLeaf, Soybean, Corn, and Grape. The matrix shows high true positive values along the diagonal, indicating correct classifications for all classes and also correct classification of healthy conditions. Misclassifications are minimal, with a few off-diagonal entries, suggesting occasional confusion between similar disease types, such as fungal diseases across crops. Healthy classes generally exhibit higher precision, reflecting the model's ability to distinguish diseased from non-diseased states effectively. The overall structure of the matrix underscores the reliability of FourCropNet in handling complex multi-class datasets, achieving high accuracy and balance across all classes. These results confirm the model's potential for practical applications in disease detection across diverse agricultural crops. To understand the computationanl requirements and its comparative study with other standard models, Table 2 shows the count of total learnable parametrers and epochs required to achieve the convergence.

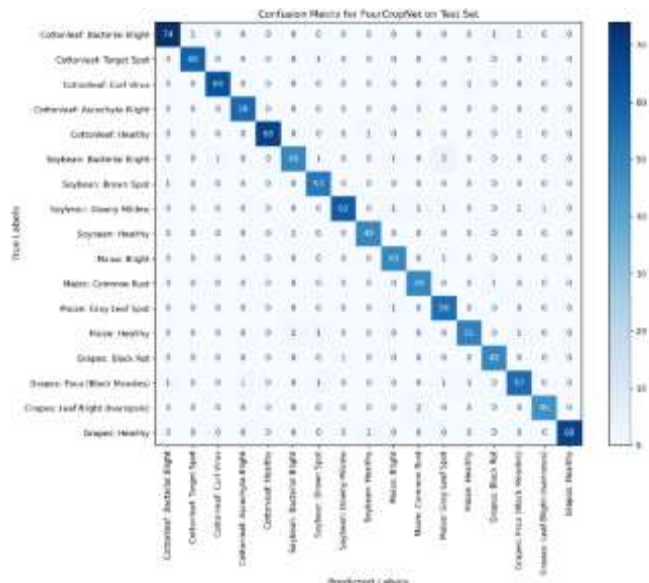

Figure 6: Confusion Matrix Analysis of FourCropNet for 15 Classes

Table 2: Comparative Model Complexities

| Model | Input Image Size | Learnable Parameters | Training Epochs |
|---|---|---|---|
| MobileNet-V3 | 224x224 | 5.4 million | 50 |
| ResNet-101 | 224x224 | 44.5 million | 100 |
| EfficientNet-B1 | 240x240 | 7.8 million | 75 |
| Inception-V3 | 299x299 | 23.8 million | 100 |
| FourCropNet (Proposed) | 224X224 | 6.5 million | 110 |

The comparative study is performed for various models across different species and datasets in terms of accuracy. For Soybean disease detection, MobileNet V2 achieved 98.61%, while the proposed FourCropNet achieved a higher accuracy of 99.5% on the Kaggle dataset [28], demonstrating its effectiveness. For Cucumber, DCCNN [29] recorded an accuracy of 98.23%. Similarly, for Grape, VGG16-based DCNN [30] achieved 99.18%, whereas FourCropNet achieved 99.7%, reflecting significant improvement.

In Corn, various YOLO models were evaluated [19], with YOLOv8n achieving the highest accuracy of 99.04%. For comparison, the proposed FourCropNet achieved 98% on Kaggle. In Tomato disease detection [31], ToLeD (a combination of VGG16, Inception, and MobileNet-V2) achieved 91.2% across nine classes. Apple disease detection results show VGG-ICNN [32] achieving 94.24%, while Res-ATTEN [33] achieved 99.00%, indicating the role of attention mechanisms in enhancing model performance.

Rice disease detection showed high performance, with ADLWNN [34] and Res-ATTEN [19] both achieving 99.00%. For Corn, the results varied, with some models achieving 94%, while multiple crops in larger datasets yielded high accuracies, such as 99.16% for 12 classes. The proposed FourCropNet demonstrated robust performance with accuracies of 96.8% for CottonLeaf, 96% for Soybean, and 95.3% for combined datasets of 15 classes.

Overall, the proposed FourCropNet consistently outperforms or matches state-of-the-art models across multiple species, showcasing its robustness, particularly in Grape and Combined crop datasets, with competitive results in Corn and CottonLeaf. These findings establish FourCropNet as a reliable and efficient solution for plant disease detection across diverse crops.

The proposed FourCropNet model demonstrates superior performance over other models due to its innovative architectural design and tailored feature extraction approach for multi-crop disease detection. Key factors contributing to its effectiveness include the integration of residual blocks for efficient feature learning, attention mechanisms to enhance focus on disease-specific regions, and the use of lightweight yet powerful layers to minimize computational overhead while maintaining accuracy. Unlike traditional models, which may struggle with overfitting or generalization issues across diverse crops, FourCropNet balances depth and efficiency to achieve robust feature representation. Its ability to adapt to varying class complexities, from single-crop datasets to combined multi-crop datasets with 15 classes, showcases its scalability. The model's high accuracy, specificity, and sensitivity across CottonLeaf, Grape, Corn and Soybean datasets reflect its capacity to capture fine-grained disease patterns. Furthermore, the use of optimized preprocessing techniques and data augmentation enhances its generalization to unseen test data, reducing the risk of performance drops in real-world scenarios. By leveraging a carefully balanced architecture with global average pooling and fully connected layers, FourCropNet ensures high classification precision and recall. These innovations collectively enable FourCropNet to outperform state-of-the-art models, making it a reliable and efficient solution for real-time multi-crop disease detection in agricultural applications.

## 4.  Conclusion

The proposed FourCropNet model demonstrates its effectiveness as a robust and scalable solution while considering its application for multi-crop disease detection, which also consideres addressing the diverse challenges posed by agricultural datasets. Through its innovative design, incorporating residual blocks for efficient feature extraction, attention mechanisms to prioritize disease-relevant regions, and a lightweight architecture, FourCropNet achieves superior accuracy alson with significant specificity for true neagaive cases and sensitivity for true positive cases, and F1 scores compared to state-of-the-art models. Its adaptability across varying class complexities, from single-crop datasets to multi-crop datasets with 15 classes, underscores its versatility and scalability. The evaluation across crops such as CottonLeaf, Grape, Potato, Soybean, and Corn highlights FourCropNet's consistent performance, achieving the highest accuracy of 99.7% for Grape and 95.3% for the combined dataset. These results validate the model's ability to be used for multiple datasets and to generalize effectively across different datasets and disease types while maintaining computational efficiency. Moreover, FourCropNet's architecture is tailored to reduce overfitting and enhance generalization, showcasing its tuitability for real-world applications. In conclusion, FourCropNet sets a new benchmark for crop disease detection models by combining cutting-edge design with practical applicability. Its consistent performance across diverse datasets positions it as a reliable tool for farmers and agricultural practitioners, paving the way for improved crop health monitoring and sustainable farming practices.